\documentclass[letterpaper, 10 pt, conference]{ieeeconf}
\usepackage{amssymb}
\usepackage{amsmath}
\usepackage{amsfonts}
\usepackage{mathrsfs }
\usepackage{stmaryrd}
\usepackage[noadjust]{cite}
\usepackage{graphicx}
\usepackage{subfigure}
\graphicspath{{pdfs/},{images/} }
\usepackage{xcolor}
\usepackage{colortbl,dcolumn}
\usepackage{algorithm}
\usepackage{algorithmic}
\usepackage{diagbox}
\usepackage{booktabs}
\usepackage{multirow}
\usepackage{verbatim}
\allowdisplaybreaks[4]

\newcommand{\figref}[1]{Fig.~\ref{#1}}
\newcommand{\tabref}[1]{Table~\ref{#1}}

\IEEEoverridecommandlockouts

\def\BibTeX{{\rm B\kern-.05em{\sc i\kern-.025em b}\kern-.08em
    T\kern-.1667em\lower.7ex\hbox{E}\kern-.125emX}}
\begin{document}

\title{\LARGE \bf
PO-MSCKF: An Efficient Visual-Inertial Odometry by Reconstructing the Multi-State Constrained Kalman Filter with the Pose-only Theory

\thanks{Authors are with the College of Intelligent Science and Technology, National University
	of Defense Technology, Changsha, 410073, China.}
\thanks{This research is funded by the National Natural Science Foundation of China (grant number: 62103430, 62103427, 62073331) and Major Project of Natural Science Foundation of Hunan Province (No. 2021JC0004)}
\thanks{$^\dag$Xueyu Du and Lilian Zhang contribute equally to this work}
\thanks{$^*$Jun Mao is the corresponding author: (maojun12@nudt.edu.cn)}
}

\author{Xueyu Du, Lilian Zhang$^\dag$, Ruochen Liu, Maosong Wang, Wenqi Wu and Jun Mao$^*$}

\maketitle

\begin{abstract}
Efficient Visual-Inertial Odometry (VIO) is crucial for payload-constrained robots. Though modern optimization-based algorithms have achieved superior accuracy, the MSCKF-based VIO algorithms are still widely demanded for their efficient and consistent performance. 
As MSCKF is built upon the conventional multi-view geometry, the measured residuals are not only related to the state errors but also related to the feature position errors. 
To apply EKF fusion, a projection process is required to remove the feature position error from the observation model, which can lead to model and accuracy degradation. 
To obtain an efficient visual-inertial fusion model, while also preserving the model consistency, we propose to reconstruct the MSCKF VIO with the novel Pose-Only (PO) multi-view geometry description. In the newly constructed filter, we have modeled PO reprojection residuals, which are solely related to the motion states and thus overcome the requirements of space projection. Moreover, the new filter does not require any feature position information, which removes the computational cost and linearization errors brought in by the 3D reconstruction procedure. We have conducted comprehensive experiments on multiple datasets, where the proposed method has shown accuracy improvements and consistent performance in challenging sequences.

\end{abstract}

\section{Introduction}\label{Introduction}
VIO technology is widely used in various mobile robots for autonomous navigation. Although different VIO algorithms have been developed in the past decade, roboticists are still expecting algorithms with lower computation costs, higher accuracy, and better robustness.

The mainstream VIO algorithms can be broadly divided into two categories: filter-based \cite{4209642, 9196524,WOS:000411059400002} and nonlinear optimization-based methods \cite{WOS:000350472800005, WOS:000712319501044, WOS:000442341000003, WOS:000725804900006,WOS:000525362000002}. 
In a VIO algorithm, IMU is used as the prediction, and visual observation is used to compensate for the errors. 
The key to fusing visual and inertial information is to build differentiable visual residuals relative to the inertial-predicted motion parameters; in multi-view geometry, the reprojection error is a golden standard to describe such residuals \cite{hartley2003multiple}.

However, the reprojection errors are related to both the camera motion and the features' 3D position, which causes most existing VIO algorithms to run pose estimation and 3D reconstruction simultaneously. As the observed feature increases, the parameters to be optimized can increase dramatically, leading to efficiency decreasing for motion estimation. Although techniques, such as sliding window optimization \cite{WOS:000280869700005}, marginalization \cite{WOS:000280869700005}, and sparsification \cite{ceres-solver,grisetti2011g2o}, have been proposed to decrease the computation burden in VO/VIO, the reconstructing of the 3D point still occupies most of the computation resources. 

\begin{figure}[htbp]
	\centering
	\includegraphics [scale=0.4]{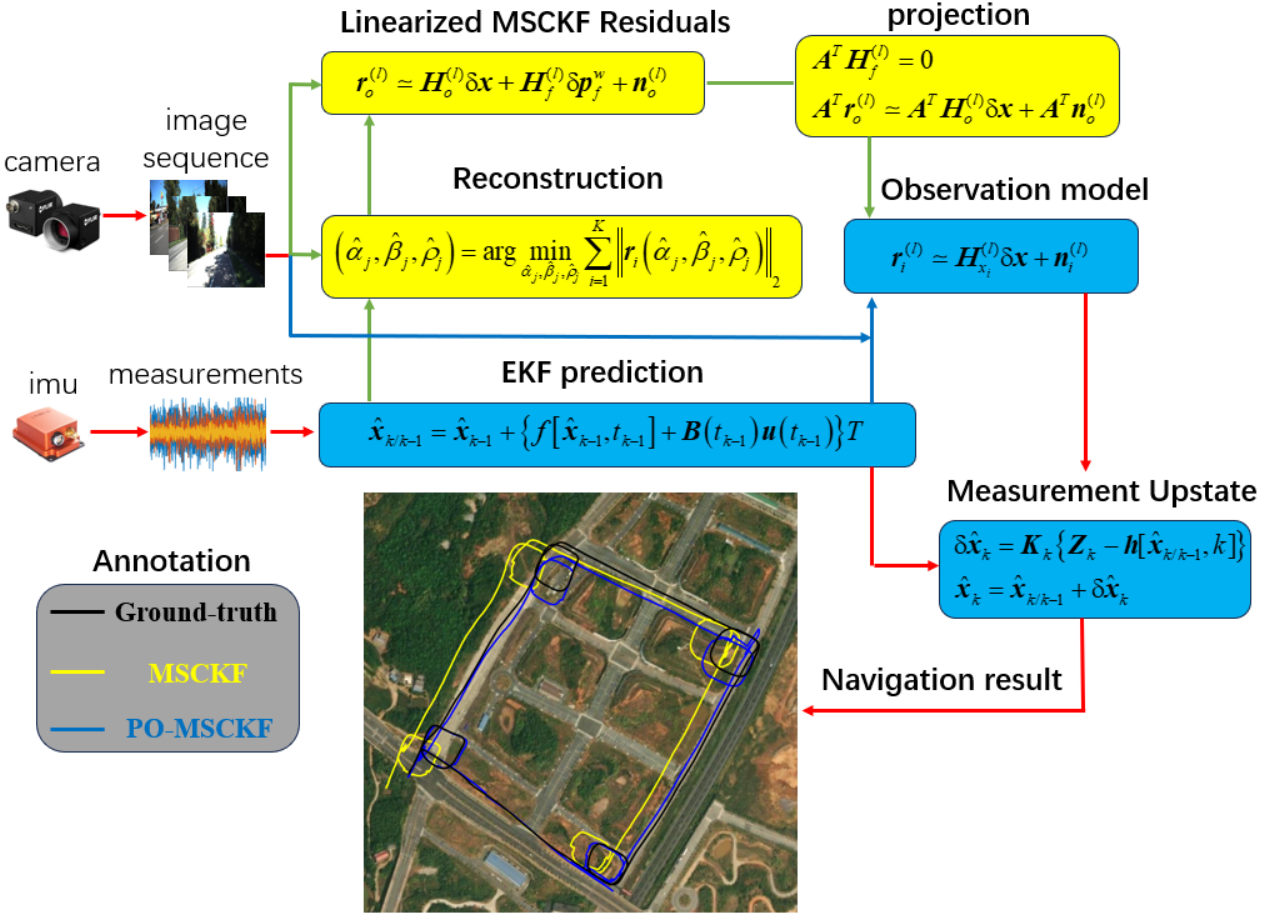}\label{fig:tech}  
	\caption{
 The complete diagram of MSCKF is shown in the figure.
 The proposed PO-MSCKF has a state space same as MSCKF to preserve the algorithm's efficiency. Meanwhile, by using the PO representation, PO-MSCKF can be processed with standard EKF (as shown by the blue boxes) and avoid the null space projection and 3D reconstruction in MSCKF (as shown by the yellow boxes). Experimental results have shown accuracy improvements on ground and aerial datasets.
}

\label{fig:tech1}  
\end{figure}

 To obtain an efficient VIO algorithm, MSCKF \cite{4209642,9196524,WOS:000320543400004,WOS:000424646100016,WOS:000570976400008,7989022} has proposed to solely maintain the camera motion parameters in the EKF state and to construct the multi-view constraints in the observation model, as shown in \figref{fig:tech1}.
 Though the measurement residuals in MSCKF are also modeled as the reprojection errors, it utilizes a null space projection operation to remove the 3D point residuals in the observation model; this helps to obtain an efficient VIO without 3D reconstructing in the filter, but the model consistency has been modified. 
 Furthermore, rooted in the limitation of the conventional multi-view geometry description, the linearization of the measurement model in MSCKF still needs the observed features' 3D position. 
 To obtain the 3D positions, MSCKF uses the predicted nominal states for efficient 3D reconstruction, but at the cost of bringing in model linearization errors caused by the inaccurate 3D estimation.

The newly proposed PO theory indicates that reprojection error can be obtained solely through camera pose, which is equivalent to traditional multi-view geometry description \cite{WOS:000458768000003,WOS:000899419900005}. Based on the PO theory, nonlinear optimization-based VO \cite{WOS:000899419900005} and VIO \cite{10438889} algorithms have been developed, and have shown improvements in accuracy and robustness.

However, benchmark comparisons have shown that, though optimization-based VIO have high precision performance, they consume more computing resources; Meanwhile, filter-based VIO requires less hardware performance and has consistent performance in different datasets \cite{WOS:000446394502008}. 
To develop efficient filter-based VIO algorithms for payload constraint robots, while also achieving notable performance in robustness and accuracy, we propose to make use of the PO theory to reconstruct the MSCKF structure. By using the PO theory, we can inherit the concise state space of MSCKF, while also avoiding the model consistency degradation caused by the measurement space projection and excluding the linearization errors caused by the inaccurate 3D reconstruction.

The main contributions are as follows:

\begin{itemize}
    \item An efficient visual-inertial EKF fusion framework is constructed, where the concise state space only includes the motion parameters and the observed residuals are linearly represented by the state errors.	
    \item The observation model in the new filter is reconstructed by using the PO multi-view geometry description, which avoids the model accuracy degradation caused by the null space projection and excludes the linearization errors brought in by the inaccurate 3D reconstruction.%

    \item Comprehensive experiments have been conducted on UAV and ground vehicle datasets, including indoor and outdoor challenging sequences.
\end{itemize}

Experimental results show the proposed method can consistently work on all the testing datasets and has shown accuracy improvements, while some algorithms failed in some challenging sequences.
   
\section{Related Work}\label{sec:Related Work}
VIO has been developed based on VO, which is proposed by Nister \cite{WOS:000223603900088} and utilizes incremental image sequences to estimate the camera pose.
The representative VO algorithms include LIBVISO2 \cite{WOS:000320772200076}, SVO \cite{WOS:000377221100003} and DSO \cite{WOS:000424465900008}.

However, due to the limitations of cameras, VO cannot work well under challenging conditions, such as complex lighting variations, motion blurring, and fast rotations.%
To improve the accuracy and robustness of the VO systems, researchers have proposed to combine inertial and visual sensors to construct a VIO system.

According to the information fusion method, VIO can be broadly divided into filter-based and optimization-based algorithms.

The early filter-based methods have been mainly developed on EKF-SLAM \cite{book}, which includes both the IMU's poses and the feature positions in the filter space for visual and inertial fusion; the disadvantage of this method is prone to the problem of feature dimension explosion \cite{WOS:000320543400004}.
As one of the most classic filter-based methods, MSCKF \cite{4209642} has proposed to remove the feature information from the filter state space and constructs the multi-view constraints in the observation model, which greatly reduces the computational complexity compared to EKF-SLAM \cite{WOS:000320543400004}.

The original MSCKF has been extended and improved in different ways, such as observability constraints \cite{7899999,WOS:000320543400004}, square-root form \cite{WOS:000570976400008, 7989022}, and the application of the Lie group theory. \cite{doi:10.1146/annurev-control-060117-105010,10179117}. Representative works of filter-based VIO also include ROVIO \cite{WOS:000411059400002} and trifocal-EKF \cite{indelman2012real}, which respectively use iterative Kalman filter and trifocal tensor to obtain accurate navigation results.

The optimization-based methods use batch graph optimization \cite{grisetti2011g2o} or bundle adjustment techniques \cite{ceres-solver} to maintain and optimize all measurements for optimal state estimates.
To achieve a constant processing time, the optimization-based VIO typically only considers the bounded sliding window of recent states as active optimization variables, while marginalizing past states and measurements \cite{WOS:000280869700005}.
For inertial constraints, the proposal of pre-integration theory \cite{WOS:000300188300006} has allowed high-rate IMU measurements to be effectively integrated into the optimization process.

Leutenegger has proposed a keyframe based VIO algorithm \cite{WOS:000350472800005}, which maintains a finite size optimization window by marginalizing old keyframes.
VINS-Mono proposed by Qin \cite{WOS:000442341000003} uses sliding windows for nonlinear optimization, which can achieve online spatial calibration and loop closures.
The current mainstream optimization-based VIO algorithms also include Kimera \cite{WOS:000712319501044}, BASALT \cite{WOS:000525362000002}, VI-DSO \cite{WOS:000446394502009}, SVO2 \cite{WOS:000399348900001}, and ORB-SLAM3\cite{WOS:000725804900006}. 
In the above VIO algorithms, EKF-SLAM \cite{book} and some optimization-based methods \cite{WOS:000350472800005,WOS:000442341000003,WOS:000712319501044,WOS:000525362000002,WOS:000725804900006} run pose estimation and 3D reconstruction simultaneously, resulting in the computational complexity increasing dramatically with the increase of the observed features.
To improve the computational efficiency, MSCKF only maintains the camera motion parameters in the system state and constructs the multi-view constraints in the observation model \cite{4209642}; however, this leads to the observation model in MSCKF violating the standard EKF model and requiring space projection for transformation, which can result in model consistency being modified.
Furthermore, as the features' 3D position is still required in the observation model, MSCKF uses the predicted nominal states for effective 3D reconstruction; but at the cost of bringing in model linearization errors caused by the inaccurate 3D estimation.
In summary, the above problems are all caused by the correlation between the reprojection errors and the features' 3D position in traditional multi-view geometry descriptions. 

To build a precise and efficient mathematical basis for large-scale visual navigation and reconstruction tasks, Cai has proposed a pose-only imaging geometry representation that is equivalent to traditional epipolar geometry and can solve the pure rotation ambiguity problem\cite{WOS:000458768000003}. Based on the new representation, they have further constructed an optimization-based visual navigation algorithm, where only the motion parameters are estimated interactively and require no nonlinear optimization for the spatial feature coordinates\cite{WOS:000899419900005}. 
Ge has implemented the PO representation to ORB-SLAM3 structure to develop an optimization-based visual-inertial algorithm. Benefiting from the consistency and efficiency of the PO representation, the PIPO-SLAM has shown accuracy and robustness improvements \cite{10438889}.

However, benchmark comparisons have shown that, though optimization-based VIO have high precision performance, they consume more computing resources; Meanwhile, filter-based VIO requires less hardware performance and has consistent performance in different datasets \cite{WOS:000446394502008}. 
Furthermore, as one of the most advanced filter-based VIO algorithms, OpenVINS can also achieve comparable accuracy to optimization-based methods \cite{9196524}.
Therefore, filter-based VIO still has high research value.

By using the PO theory \cite{WOS:000458768000003,WOS:000899419900005}, we can inherit the concise state space of MSCKF, while also
avoiding the model consistency degradation caused by the measurement space projection and excluding the linearization errors caused by the inaccurate 3D reconstruction; this will help develop an efficient and robust VIO algorithm suitable for payload constraint robots.

\section{Method}\label{sec:Method}
This section presents the method of the proposed PO-MSCKF, including the foundation of the PO theory, the system model, and the observation model of our algorithm.

\subsection{Fundamentals of Pose-only Theory}\label{sec:Fundamentals of Pose-only Theory}

Before delving into the details of the proposed method, we firstly introduce the theoretical foundation of the PO theory. 
\subsubsection{Pose-only descriptions of multiple view geometry}\label{sec:Pose-only descriptions of multiple view geometry}
Assuming a 3D feature $\boldsymbol{p}_{f}^{w}=\begin{bmatrix} {X}_{f}^{w} & {Y}_{f}^{w} & {Z}_{f}^{w} \end{bmatrix}$ observed in $n$ images, its normalized coordinate in the $i$-th image is $\boldsymbol{p}_{{c}_{i}}=\begin{bmatrix} {X}_{{c}_{i}} & {Y}_{{c}_{i}} & 1 \end{bmatrix}$ ($i=1,...n$). 

Defining the global rotation and global translation of camera in the $i$-th image as $\boldsymbol{R}_{w}^{{c}_{i}}$ and $\boldsymbol{t}_{{c}_{i}}^{w}$, then the projection equation of 3D feature $\boldsymbol{p}_{f}^{w}$ can be expressed as:

\begin{align}\label{eqn:feature projection}	
		\boldsymbol{p}_{{c}_{i}}=\frac{1}{{Z}_{f}^{{c}_{i}}}{p}_{f}^{{c}_{i}}=\frac{1}{{Z}_{f}^{{c}_{i}}}\boldsymbol{R}_{w}^{{c}_{i}}(\boldsymbol{p}_{f}^{w}-\boldsymbol{t}_{{c}_{i}}^{w})
\end{align}
where $\boldsymbol{p}_{f}^{{c}_{i}}=\begin{bmatrix} {X}_{f}^{{c}_{i}} & {Y}_{f}^{{c}_{i}} & {Z}_{f}^{{c}_{i}} \end{bmatrix}$ represents the position of 3D feature $\boldsymbol{p}_{f}^{w}$ in the $i$-th camera frame and 
${Z}_{f}^{{c}_{i}} \textgreater 0$ represents the depth of this feature.

For a pair of views composed of the $i$-th and $j$-th images, their projection relationship satisfies:

\begin{align}\label{eqn:feature projection ij}	
	{Z}_{f}^{{c}_{j}}\boldsymbol{p}_{{c}_{j}}={Z}_{f}^{{c}_{i}}\boldsymbol{R}_{{c}_{i}}^{{c}_{j}}\boldsymbol{p}_{{c}_{i}}+\boldsymbol{t}_{{c}_{i}}^{{c}_{j}}
\end{align}
where $\boldsymbol{R}_{{c}_{i}}^{{c}_{j}}=\boldsymbol{R}_{w}^{{c}_{j}}\boldsymbol{R}_{{c}_{i}}^{w}$ represents the relative rotation between two images and $\boldsymbol{t}_{{c}_{i}}^{{c}_{j}}=\boldsymbol{R}_{w}^{{c}_{j}}(\boldsymbol{t}_{{c}_{i}}^{w}-\boldsymbol{t}_{{c}_{j}}^{w})$ represents the relative translation between two images. 

Left multiply the anti-symmetric matrix $[\boldsymbol{p}_{{c}_{j}}\times]$ on both sides of \eqref{eqn:feature projection ij}:
\begin{align}\label{eqn:feature projection anti}	
	{Z}_{f}^{{c}_{i}}[\boldsymbol{p}_{{c}_{j}}\times]\boldsymbol{R}_{{c}_{i}}^{{c}_{j}}\boldsymbol{p}_{{c}_{i}}=-[\boldsymbol{p}_{{c}_{j}}\times]\boldsymbol{t}_{{c}_{i}}^{{c}_{j}}
\end{align}
where ${Z}_{f}^{{c}_{i}}$ is constant and taking the magnitude, \eqref{eqn:feature projection anti} can be expressed as:

\begin{align}\label{eqn:depth i}	
	{Z}_{f}^{{c}_{i}}=\frac{||[\boldsymbol{p}_{{c}_{j}}\times]\boldsymbol{t}_{{c}_{i}}^{{c}_{j}}||}{{\boldsymbol{\theta}}_{(i,j)}} \triangleq {d}_{i}^{(i,j)}
\end{align}
where ${\boldsymbol{\theta}}_{(i,j)}=||[\boldsymbol{p}_{{c}_{j}}\times]\boldsymbol{R}_{{c}_{i}}^{{c}_{j}}\boldsymbol{p}_{{c}_{i}}||$. 
Similarly to \eqref{eqn:feature projection anti}, left multiply the anti-symmetric matrix $[(\boldsymbol{R}_{{c}_{i}}^{{c}_{j}}\boldsymbol{p}_{{c}_{i}})\times]$ on both sides of \eqref{eqn:feature projection ij}:

\begin{align}\label{eqn:depth j}	
	{Z}_{f}^{{c}_{j}}=\frac{||[(\boldsymbol{R}_{{c}_{i}}^{{c}_{j}}\boldsymbol{p}_{{c}_{i}})\times]\boldsymbol{t}_{{c}_{i}}^{{c}_{j}}||}{{\boldsymbol{\theta}}_{(i,j)}} \triangleq {d}_{j}^{(i,j)}
\end{align}

By combining \eqref{eqn:depth i} and \eqref{eqn:depth j}, the pose-only constraint for the two-view geometry can be obtained, named a pair of pose-only (PPO) constraints \cite{WOS:000458768000003}:
\begin{align}\label{eqn:depth ij}	
	{d}_{j}^{(i,j)}\boldsymbol{p}_{{c}_{j}}={d}_{i}^{(i,j)}\boldsymbol{R}_{{c}_{i}}^{{c}_{j}}\boldsymbol{p}_{{c}_{i}}+\boldsymbol{t}_{{c}_{i}}^{{c}_{j}}
\end{align}

In \eqref{eqn:depth ij}, the depth information of 3D features is represented by relative poses and 2D features, meaning that PPO constraints are equivalent to traditional two-view geometry without the need for 3D features \cite{WOS:000458768000003}.

Based on PPO constraints, \cite{WOS:000899419900005} has further derived the pose-only multi-view geometry description: 
\begin{align}\label{eqn:constraint mulimage}	
	D(j,k)=\left\{{d}_{i}^{(j,i)}\boldsymbol{p}_{{c}_{i}}={d}_{j}^{(j,k)}\boldsymbol{R}_{{c}_{j}}^{{c}_{i}}\boldsymbol{p}_{{c}_{j}}+\boldsymbol{t}_{{c}_{j}}^{{c}_{i}}|1 \leq i \leq n, i \neq j\right\}
\end{align}

where the $j$-th and $k$-th images represent the left and right base views of the constraint. The suggested base view selection method can be expressed as \cite{WOS:000899419900005}:
\begin{align}\label{eqn:baseframe sel}	
	(j,k)=\mathop{argmax}\limits_{1 \leq jj,kk \leq n} \left\{ \boldsymbol{\theta}_{jj,kk} \right\} 
\end{align}

Equation \eqref{eqn:constraint mulimage} means that when a 3D feature is observed in $n$ images, traditional multi-view geometry descriptions can be equivalently expressed in a pose-only form \cite{WOS:000899419900005}. 

\subsubsection{Reprojection model based on pose-only constraints}\label{sec:Reprojection model based on pose-only constraints}
In traditional multi-view geometry descriptions, assuming $\boldsymbol{p}_{{c}_{i}}^{(l)}$ is the normalized coordinate of feature $l$ in $i$-th image, the reprojection error can be defined as:

\begin{align}\label{eqn:repro BA}
	\boldsymbol{r}_{{c}_{i}}^{(l)}=\boldsymbol{\widetilde{p}}_{{c}_{i}}^{(l)}-\boldsymbol{p}_{{c}_{i}}^{(l)}=\frac{\boldsymbol{p}_{{f}_{l}}^{{c}_{i}(BA)}}{{{e}_{3}^{T}}\boldsymbol{p}_{{f}_{l}}^{{c}_{i}(BA)}}-\boldsymbol{p}_{{c}_{i}}^{(l)}
\end{align}
where $\boldsymbol{p}_{{f}_{l}}^{{c}_{i}(BA)}=\boldsymbol{R}_{w}^{{c}_{i}}(\boldsymbol{p}_{{f}_{l}}^{{w}(BA)}-\boldsymbol{t}_{{c}_{i}}^{w})$ and ${e}_{3}^{T}=\begin{bmatrix} 0 & 0 & 1 \end{bmatrix}$. 

The 3D feature position $\boldsymbol{p}_{{f}_{l}}^{{w}(BA)}$ can be estimated through a nonlinear optimization process that minimizes reprojection errors, mainly achieved through triangulation measurement and Bundle Adjustment (BA) \cite{10.5555/1888028.1888032}.

After using PO multi-view geometry description in \eqref{eqn:constraint mulimage}, the reprojection error can be redefined as:

\begin{align}\label{eqn:repro PA}
	\boldsymbol{r}_{{c}_{i}}^{(l)}=\boldsymbol{\widetilde{p}}_{{c}_{i}}^{(l)}-\boldsymbol{p}_{{c}_{i}}^{(l)}=\frac{\boldsymbol{p}_{{f}_{l}}^{{c}_{i}(PO)}}{{{e}_{3}^{T}}\boldsymbol{p}_{{f}_{l}}^{{c}_{i}(PO)}}-\boldsymbol{p}_{{c}_{i}}^{(l)}
\end{align}

where $\boldsymbol{p}_{{f}_{l}}^{{c}_{i}(PO)}$ can be directly obtained through the relative camera pose and 2D features as:

\begin{align}\label{eqn:3Dpos PO}
	\boldsymbol{p}_{{f}_{l}}^{{c}_{i}(PO)}= ||[{\boldsymbol{t}_{{c}_{j}}^{{c}_{k}}}\times]\boldsymbol{p}_{{c}_{k}}||\boldsymbol{R}_{{c}_{j}}^{{c}_{i}}\boldsymbol{p}_{{c}_{j}}+||[{\boldsymbol{p}_{{c}_{k}}}\times]\boldsymbol{R}_{{c}_{j}}^{{c}_{k}}{\boldsymbol{p}_{{c}_{j}}}||{\boldsymbol{t}_{{c}_{j}}^{{c}_{i}}}
\end{align}
where the $j$-th and $k$-th images represent the left and right base views respectively.

Compared to \eqref{eqn:repro BA}, the reprojection error defined in \eqref{eqn:repro PA} is independent of 3D feature position, eliminating the need for the 3D reconstruction process.

\subsection{System Model}\label{sec:System Model}

PO-MSCKF adopts a system model consistent with MSCKF \cite{4209642}, and the state vector is composed of the IMU motion state and camera pose:
\begin{align}\label{eqn:x PO}
	\boldsymbol{x}=\begin{bmatrix} \boldsymbol{x}_{b} & \boldsymbol{x}_{c} \end{bmatrix} ^{T}
\end{align}
where $\boldsymbol{x}_{b}$ is the state vector related to the IMU, and $\boldsymbol{x}_{c}$ is the state vector related to the camera.

$\boldsymbol{x}_{b}$ includes the IMU attitude, velocity, position, gyroscope biases, and accelerometer biases, defined as follows:
\begin{align}\label{eqn:xb PO}
	\boldsymbol{x}_{b}=\begin{bmatrix} \boldsymbol{q}_{b}^{w} & \boldsymbol{v}_{b}^{w} & \boldsymbol{p}_{b}^{w} & \boldsymbol{b}_{g} & \boldsymbol{b}_{a} \end{bmatrix} ^{T}
\end{align}

$\boldsymbol{x}_{c}$ includes the camera pose of the past $N$ frames' position, where $N$ depends on the number of the tracked features and the maximum length of the sliding window. $\boldsymbol{x}_{c}$ can be defined as follows:
\begin{align}\label{eqn:xc PO}
	\boldsymbol{x}_{c}=\begin{bmatrix} \boldsymbol{q}_{{c}_{1}}^{w} & \boldsymbol{p}_{{c}_{1}}^{w} & {...} & \boldsymbol{q}_{{c}_{N}}^{w} & \boldsymbol{p}_{{c}_{N}}^{w} \end{bmatrix} ^{T}
\end{align}

For velocity, position, and biases, the error definition is: 
$\delta \boldsymbol{x}=\boldsymbol{\widetilde{x}}-\boldsymbol{x} $ 
where $\boldsymbol{x}$ is the true value and $\boldsymbol{\widetilde{x}}$ is the estimated value. The orientation error is defined through the difference between the true and the estimated quaternion as:
\begin{align}\label{eqn:quaternions}
	\boldsymbol{q}=\delta \boldsymbol{q} \otimes \boldsymbol{\widetilde{q}}\approx\begin{bmatrix}
	    1 \\ \frac{1}{2} \boldsymbol {\phi} 
	\end{bmatrix} \otimes  \boldsymbol{\widetilde{q}}
\end{align} 
where $\otimes$ represents quaternion multiplication.

The corresponding error state vector is:
\begin{align}\label{eqn:deltax PO}
	\delta \boldsymbol{x}=\begin{bmatrix} \delta \boldsymbol{x}_{b} & \delta \boldsymbol{x}_{c} \end{bmatrix} ^{T}
\end{align}

$\delta \boldsymbol{x}_{b}$ include the IMU attitude error, velocity error, position error, gyroscope biases error, and accelerometer biases error, defined as follows:
\begin{align}\label{eqn:deltaxb PO}
	\delta \boldsymbol{x}_{b}=\begin{bmatrix} \boldsymbol{\phi}_{b}^{w} & \delta \boldsymbol{v}_{b}^{w} & \delta \boldsymbol{p}_{b}^{w} & \delta \boldsymbol{b}_{g} & \delta \boldsymbol{b}_{a} \end{bmatrix} ^{T}
\end{align}

The system error state model of IMU can be written: 

\begin{align}\label{eqn:deltaxb process PO}
	\delta \boldsymbol{\dot{x}}_{b}&=\boldsymbol{F}_{b} \delta \boldsymbol{x}_{b}+\boldsymbol{G}_{b}\boldsymbol{w}_{b} \\ \boldsymbol{w}_{b}&=\begin{bmatrix} \boldsymbol{w}_{g} & \boldsymbol{w}_{a} & \boldsymbol{w}_{wg} & \boldsymbol{w}_{wa} \end{bmatrix} ^{T}
\end{align}
where $\boldsymbol{w}_{b}$ is the system noise vector.  
$\boldsymbol{w}_{g}$ and $\boldsymbol{w}_{a}$ represent the measurement white noise of the gyro and accelerometer respectively; $\boldsymbol{w}_{wg}$ and $\boldsymbol{w}_{wa}$ represent the driven white noise of the gyro biases and accelerometer biases.

According to \cite{4209642}, the error-state transition matrix $\boldsymbol{F}_{b}$ and the input noise Jacobian matrix $\boldsymbol{G}_{b}$ can be represented as:  
\begin{align}\label{eqn:F PO}
	\boldsymbol{F}_{b}=
	\begin{bmatrix}
		- [\boldsymbol{\omega}_{ie}^{w}\times] & \boldsymbol{0}_{3\times3} & \boldsymbol{0}_{3\times3} & -\boldsymbol{R}_{b}^{w} & \boldsymbol{0}_{3\times3} \\
		[(\boldsymbol{R}_{b}^{w}\boldsymbol{f}^{b})\times] & -2 [\boldsymbol{\omega}_{ie}^{w}\times] & \boldsymbol{0}_{3\times3} & \boldsymbol{0}_{3\times3} & \boldsymbol{R}_{b}^{w} \\
		\boldsymbol{0}_{3\times3} & \boldsymbol{I}_{3\times3} & \boldsymbol{0}_{3\times3} & \boldsymbol{0}_{3\times3} &\boldsymbol{0}_{3\times3} \\
		\boldsymbol{0}_{3\times3} & \boldsymbol{0}_{3\times3} & \boldsymbol{0}_{3\times3} & \boldsymbol{0}_{3\times3} &\boldsymbol{0}_{3\times3} \\
		\boldsymbol{0}_{3\times3} & \boldsymbol{0}_{3\times3} & \boldsymbol{0}_{3\times3} & \boldsymbol{0}_{3\times3} &\boldsymbol{0}_{3\times3}
	\end{bmatrix}
\end{align}
\begin{align}\label{eqn:G PO}
	\boldsymbol{G}_{b}=
	\begin{bmatrix}
		-\boldsymbol{R}_{b}^{w} & \boldsymbol{0}_{3\times3} & \boldsymbol{0}_{3\times3} & \boldsymbol{0}_{3\times3} \\
		\boldsymbol{0}_{3\times3} & \boldsymbol{R}_{b}^{w} & \boldsymbol{0}_{3\times3} & \boldsymbol{0}_{3\times3} \\
		\boldsymbol{0}_{3\times3} & \boldsymbol{0}_{3\times3} & \boldsymbol{0}_{3\times3} & \boldsymbol{0}_{3\times3} \\
		\boldsymbol{0}_{3\times3} & \boldsymbol{0}_{3\times3} & \boldsymbol{I}_{3\times3} & \boldsymbol{0}_{3\times3} \\
		\boldsymbol{0}_{3\times3} & \boldsymbol{0}_{3\times3} & \boldsymbol{0}_{3\times3} & \boldsymbol{I}_{3\times3} 
	\end{bmatrix}
\end{align}
where $\boldsymbol{\omega}_{ie}^{w}$ is the projection of the earth's rotational velocity in the $w$ system, $\boldsymbol{f}^{b}$ represents the specific force measurement of IMU.

The error state $\delta \boldsymbol{x}_{c}$ can be represented as:
\begin{align}\label{eqn:deltaxc PO}
	\delta \boldsymbol{x}_{c}=\begin{bmatrix} \boldsymbol{\phi}_{{c}_{1}}^{w} & \delta \boldsymbol{p}_{{c}_{1}}^{w} & {...} & \boldsymbol{\phi}_{{c}_{N}}^{w} & \delta \boldsymbol{p}_{{c}_{N}}^{w} \end{bmatrix} ^{T}
\end{align}

After obtaining the measurement for each key frame, a new camera pose state is added to the state vector, and the state covariance matrix is augmented. 
In state augmentation, the new camera error state can be represented as:

\begin{align}\label{eqn:deltaxc add PO}
	\delta \boldsymbol{x}_{{c}_{N+1}}=\begin{bmatrix} \boldsymbol{\phi}_{{c}_{N+1}}^{w} & \delta \boldsymbol{p}_{{c}_{N+1}}^{w} \end{bmatrix} ^{T} \approx \boldsymbol{J} \delta \boldsymbol{x}
\end{align}
where $J$ represents the Jacobian matrix between $\delta \boldsymbol{x}_{{c}_{N+1}}$ and $\delta \boldsymbol{x}$ \cite{4209642}:
\begin{align}\label{eqn:J PO}
	\boldsymbol{J}=
	\begin{bmatrix}
		\boldsymbol{I}_{3\times3} & \boldsymbol{0}_{3\times3} & \boldsymbol{0}_{3\times3} & \boldsymbol{0}_{3\times6} & \boldsymbol{0}_{3\times6N} \\
		[(\boldsymbol{R}_{b}^{w}\boldsymbol{p}_{c}^{b})\times] & \boldsymbol{0}_{3\times3} & \boldsymbol{I}_{3\times3} & \boldsymbol{0}_{3\times6} & \boldsymbol{0}_{3\times6N}
	\end{bmatrix}
\end{align}

The augmented covariance matrix can be expressed as \cite{4209642}:
\begin{align}\label{eqn:augmentation P}
	\boldsymbol{P}_{k}^{*} &=
	\begin{bmatrix}
		\boldsymbol{I}_{6N+15}  \\
		\boldsymbol{J}
	\end{bmatrix}
	\boldsymbol{P}_{k}
		\begin{bmatrix}
		\boldsymbol{I}_{6N+15} & \boldsymbol{J}^{T}
	\end{bmatrix}
	\\
	&=
	\begin{bmatrix}
		\boldsymbol{P}_{k} & \boldsymbol{P}_{(bc){{c}_{N+1}}}  
		\\
		\boldsymbol{P}_{(bc){{c}_{N+1}}}^{T} & \boldsymbol{P}_{{{c}_{N+1}}{{c}_{N+1}}} 
	\end{bmatrix} \nonumber
\end{align}
where $\boldsymbol{P}_{k}$ and $\boldsymbol{P}_{k}^{*}$ represent the covariance matrices before and after augmentation respectively. $\boldsymbol{P}_{(bc){{c}_{N+1}}}$ and $\boldsymbol{P}_{{{c}_{N+1}}{{c}_{N+1}}}$ are defined as \cite{4209642}:
\begin{align}\label{eqn:augmentation Pbc}
	\boldsymbol{P}_{(bc){{c}_{N+1}}} &=E(\delta \boldsymbol{x} \delta \boldsymbol{x}_{{c}_{N+1}}^{T})
	=E(\delta \boldsymbol{x} \delta\boldsymbol{x}^{T} \boldsymbol{J}^{T})=\boldsymbol{P}_{k}\boldsymbol{J}^{T} \nonumber
	\\
	\boldsymbol{P}_{{{c}_{N+1}}{{c}_{N+1}}} &=E(\delta \boldsymbol{x}{{c}_{N+1}} \delta \boldsymbol{x}_{{c}_{N+1}}^{T})=E(\boldsymbol{J} \delta \boldsymbol{x} \delta\boldsymbol{x}^{T} \boldsymbol{J}^{T})=\boldsymbol{J}\boldsymbol{P}_{k}\boldsymbol{J}^{T} 
\end{align}

\subsection{Observation Model}\label{sec:Measurement Model}

In MSCKF, the reprojection error is related to the camera poses and 3D features, but the system state does not contain 3D features; therefore, it is first represented by the system error state and 3D features error, and then restore the standard EKF format using a null space projection \cite{4209642}. 

In \eqref{eqn:repro PA}, the reprojection error $\boldsymbol{r}_{{c}_{i}}^{(l)}$ is independent of 3D features, then it can be rewritten as:

\begin{align}\label{eqn:reproject PO}
	\boldsymbol{r}_{{c}_{i}}^{(l)} \approx \boldsymbol{H}_{{x}_{i}}^{(l)} \delta \boldsymbol{x} + \boldsymbol{n}_{i}^{(l)}
\end{align}

Compared to MSCKF \cite{4209642}, \eqref{eqn:reproject PO} is the standard EKF observation model and therefore does not require a null space projection.

$\boldsymbol{H}_{{x}_{i}}^{(l)}$ is the Jacobian matrix of $\boldsymbol{r}_{{c}_{i}}^{(l)}$ relative to $\delta \boldsymbol{x}$, and the derivation process is as follows:

\begin{equation}\label{eqn:Hxi PO}
	\begin{aligned}
		\boldsymbol{H}_{{x}_{i}}^{(l)}&=\frac{\partial\boldsymbol{r}_{{c}_{i}}^{(l)}}{\partial\delta \boldsymbol{x}}
		=\frac{\partial\boldsymbol{r}_{{c}_{i}}^{(l)}}{\partial\boldsymbol{p}_{{f}_{l}}^{{{c}_{i}}(PO)}}
		\frac{\partial\boldsymbol{p}_{{f}_{l}}^{{{c}_{i}}(PO)}}{\partial\delta \boldsymbol{x}} \\
		&=\begin{bmatrix}
			\frac{1}{{e}_{3}^{T}\boldsymbol{p}_{{f}_{l}}^{{{c}_{i}}(PO)}} & 0 & -\frac{{e}_{1}^{T}\boldsymbol{p}_{{f}_{l}}^{{{c}_{i}}(PO)}}{{({e}_{3}^{T}\boldsymbol{p}_{{f}_{l}}^{{{c}_{i}}(PO)})}^2} \\
			0 & \frac{1}{{e}_{3}^{T}\boldsymbol{p}_{{f}_{l}}^{{{c}_{i}}(PO)}} & -\frac{{e}_{2}^{T}\boldsymbol{p}_{{f}_{l}}^{{{c}_{i}}(PO)}}{{({e}_{3}^{T}\boldsymbol{p}_{{f}_{l}}^{{{c}_{i}}(PO)})}^2}
		\end{bmatrix} \\
		&~~\cdot\begin{bmatrix}
			\boldsymbol{0}_{3\times15} & \frac{\partial\boldsymbol{p}_{{f}_{l}}^{{{c}_{i}}(PO)}}{\partial\boldsymbol{\phi}_{{c}_{ii}}^{w}} & \frac{\partial\boldsymbol{p}_{{f}_{l}}^{{{c}_{i}}(PO)}}{\partial\delta\boldsymbol{p}_{{c}_{ii}}^{w}} & ...
		\end{bmatrix}
	\end{aligned}
\end{equation}
where ${e}_{1}^{T}=\begin{bmatrix} 1 & 0 & 0 \end{bmatrix}$, ${e}_{2}^{T}=\begin{bmatrix} 0 & 1 & 0 \end{bmatrix}$, and ${e}_{3}^{T}=\begin{bmatrix} 0 & 0 & 1 \end{bmatrix}$.

The partial derivative of $\boldsymbol{p}_{{f}_{l}}^{{{c}_{i}}(PO)}$ with respect to $\boldsymbol{\phi}_{{c}_{ii}}^{w}$ can be expressed as:

\begin{equation}\label{eqn:Hxi phi PO}
	\begin{aligned}
		\frac{\partial\boldsymbol{p}_{{f}_{l}}^{{{c}_{i}}(PO)}}{\partial\boldsymbol{\phi}_{{c}_{ii}}^{w}}&=
		\frac{\partial(||[\boldsymbol{t}_{{c}_{j}}^{{c}_{k}}\times]\boldsymbol{p}_{{c}_{k}}||
			\boldsymbol{R}_{{c}_{j}}^{{c}_{i}}\boldsymbol{p}_{{c}_{j}}
			+||[\boldsymbol{p}_{{c}_{k}}\times]\boldsymbol{R}_{{c}_{j}}^{{c}_{k}}\boldsymbol{p}_{{c}_{j}}||\boldsymbol{t}_{{c}_{j}}^{{c}_{i}})}{\partial\boldsymbol{\phi}_{{c}_{ii}}^{w}} \\
			&=A+B+C+D
	\end{aligned}
\end{equation}

Each part in \eqref{eqn:Hxi phi PO} can be specifically represented as follows:
\begin{equation}\label{eqn:Hxi phi PO A}
	\begin{aligned}
		A&=\frac{\partial(||[\boldsymbol{t}_{{c}_{j}}^{{c}_{k}}\times]\boldsymbol{p}_{{c}_{k}}||
			\boldsymbol{R}_{{c}_{j}}^{{c}_{i}}\boldsymbol{p}_{{c}_{j}})}{\partial(||[\boldsymbol{t}_{{c}_{j}}^{{c}_{k}}\times]\boldsymbol{p}_{{c}_{k}}||)}
		\cdot \frac{\partial(||[\boldsymbol{t}_{{c}_{j}}^{{c}_{k}}\times]\boldsymbol{p}_{{c}_{k}}||)}{\partial\boldsymbol{\phi}_{{c}_{ii}}^{w}} \\
		&=\boldsymbol{R}_{{c}_{j}}^{{c}_{i}}\boldsymbol{p}_{{c}_{j}}{{A}_{1}}
	\end{aligned}
\end{equation}

\begin{equation}\label{eqn:Hxi phi PO A1}
	\begin{aligned}
		{{A}_{1}}&=\frac{\partial(||[\boldsymbol{t}_{{c}_{j}}^{{c}_{k}}\times]\boldsymbol{p}_{{c}_{k}}||)}{\partial\boldsymbol{\phi}_{{c}_{ii}}^{w}}
		=\frac{\partial(||[\boldsymbol{t}_{{c}_{j}}^{{c}_{k}}\times]\boldsymbol{p}_{{c}_{k}}||)}{\partial\boldsymbol{t}_{{c}_{j}}^{{c}_{k}}} \cdot 
		\frac{\partial\boldsymbol{t}_{{c}_{j}}^{{c}_{k}}}{\partial\boldsymbol{\phi}_{{c}_{ii}}^{w}} \\
		&=\begin{cases}
			ii=k:-(\frac{{p}_{{c}_{k}}^{T}{p}_{{c}_{k}}(\boldsymbol{t}_{{c}_{k}})^{T}-{p}_{{c}_{k}}^{T}\boldsymbol{t}_{{c}_{k}}{p}_{{c}_{k}}^{T}}{||[\boldsymbol{t}_{{c}_{j}}^{{c}_{k}}\times]\boldsymbol{p}_{{c}_{k}}||}) 
			\cdot [\boldsymbol{t}_{{c}_{j}}^{{c}_{k}}\times] \boldsymbol{R}_{w}^{{c}_{k}}\\
			ii=others:0
		\end{cases}
	\end{aligned}
\end{equation}

\begin{equation}\label{eqn:Hxi phi PO B}
	\begin{aligned}
		B&=(||[\boldsymbol{t}_{{c}_{j}}^{{c}_{k}}\times]\boldsymbol{p}_{{c}_{k}}||)\cdot
		\frac{\partial(\boldsymbol{R}_{{c}_{j}}^{{c}_{i}}\boldsymbol{p}_{{c}_{j}})}{\partial\boldsymbol{\phi}_{{c}_{ii}}^{w}} \\
		&= \begin{cases}
			i=j,ii=any:0 \\
			i \neq j:\begin{cases}
						ii=j:(||[\boldsymbol{t}_{{c}_{j}}^{{c}_{k}}\times]\boldsymbol{p}_{{c}_{k}}||)\boldsymbol{R}_{w}^{{c}_{i}}[{(\boldsymbol{R}_{{c}_{j}}^{w}\boldsymbol{p}_{{c}_{j}})\times]} \\
						ii=i:-(||[\boldsymbol{t}_{{c}_{j}}^{{c}_{k}}\times]\boldsymbol{p}_{{c}_{k}}||)\boldsymbol{R}_{w}^{{c}_{i}}[{(\boldsymbol{R}_{{c}_{j}}^{w}\boldsymbol{p}_{{c}_{j}})\times]} \\
						ii=others:0
					 \end{cases}
		   \end{cases}
	\end{aligned}
\end{equation}

\begin{equation}\label{eqn:Hxi phi PO C}
	\begin{aligned}
	C&=\frac{\partial(||[\boldsymbol{p}_{{c}_{k}}\times]\boldsymbol{R}_{{c}_{j}}^{{c}_{k}}\boldsymbol{p}_{{c}_{j}}||\boldsymbol{t}_{{c}_{j}}^{{c}_{i}})}{\partial(||[\boldsymbol{p}_{{c}_{k}}\times]\boldsymbol{R}_{{c}_{j}}^{{c}_{k}}\boldsymbol{p}_{{c}_{j}}||)}
	\cdot \frac{\partial(||[\boldsymbol{p}_{{c}_{k}}\times]\boldsymbol{R}_{{c}_{j}}^{{c}_{k}}\boldsymbol{p}_{{c}_{j}}||)}{\partial\boldsymbol{\phi}_{{c}_{ii}}^{w}} \\
	&=0
	\end{aligned}
\end{equation}

\begin{equation}\label{eqn:Hxi phi PO D}
	\begin{aligned}
		D&=(||[\boldsymbol{p}_{{c}_{k}}\times]\boldsymbol{R}_{{c}_{j}}^{{c}_{k}}\boldsymbol{p}_{{c}_{j}}||)
		\cdot \frac{\partial(\boldsymbol{t}_{{c}_{j}}^{{c}_{i}})}{\partial\boldsymbol{\phi}_{{c}_{ii}}^{w}} \\
		&=\begin{cases}
			i=j,ii=any:0 \\
			i \neq j:\begin{cases}
						ii=i:-(||[\boldsymbol{p}_{{c}_{k}}\times]\boldsymbol{R}_{{c}_{j}}^{{c}_{k}}\boldsymbol{p}_{{c}_{j}}||) [\boldsymbol{t}_{{c}_{j}}^{{c}_{i}}\times]\boldsymbol{R}_{w}^{{c}_{i}} \\
						ii=others:0
					 \end{cases}
		  \end{cases}
	\end{aligned}
\end{equation}

The partial derivative of $\boldsymbol{p}_{{f}_{l}}^{{{c}_{i}}(PO)}$ with respect to $\delta\boldsymbol{p}_{{c}_{ii}}^{w}$ can be expressed as:

\begin{equation}\label{eqn:Hxi pos PO}
	\begin{aligned}
		\frac{\partial\boldsymbol{p}_{{f}_{l}}^{{{c}_{i}}(PO)}}{\partial\delta\boldsymbol{p}_{{c}_{ii}}^{w}}&=
		\frac{\partial(||[\boldsymbol{t}_{{c}_{j}}^{{c}_{k}}\times]\boldsymbol{p}_{{c}_{k}}||
			\boldsymbol{R}_{{c}_{j}}^{{c}_{i}}\boldsymbol{p}_{{c}_{j}}
			+||[\boldsymbol{p}_{{c}_{k}}\times]\boldsymbol{R}_{{c}_{j}}^{{c}_{k}}\boldsymbol{p}_{{c}_{j}}||\boldsymbol{t}_{{c}_{j}}^{{c}_{i}})}{\partial\delta\boldsymbol{p}_{{c}_{ii}}^{w}} \\
		&=E+F
	\end{aligned}
\end{equation}
Each part in \eqref{eqn:Hxi pos PO} can be specifically represented as follows:
\begin{equation}\label{eqn:Hxi pos PO E}
	\begin{aligned}
		E&=\frac{\partial(||[\boldsymbol{t}_{{c}_{j}}^{{c}_{k}}\times]\boldsymbol{p}_{{c}_{k}}||
			\boldsymbol{R}_{{c}_{j}}^{{c}_{i}}\boldsymbol{p}_{{c}_{j}})}{\partial(||[\boldsymbol{t}_{{c}_{j}}^{{c}_{k}}\times]\boldsymbol{p}_{{c}_{k}}||)}
		\cdot \frac{\partial(||[\boldsymbol{t}_{{c}_{j}}^{{c}_{k}}\times]\boldsymbol{p}_{{c}_{k}}||)}{\partial\delta\boldsymbol{p}_{{c}_{ii}}^{w}} \\
		&=\boldsymbol{R}_{{c}_{j}}^{{c}_{i}}\boldsymbol{p}_{{c}_{j}}{{E}_{1}}
	\end{aligned}
\end{equation}

\begin{equation}\label{eqn:Hxi pos PO E1}
	\begin{aligned}
		{{E}_{1}}&=\frac{\partial(||[\boldsymbol{t}_{{c}_{j}}^{{c}_{k}}\times]\boldsymbol{p}_{{c}_{k}}||)}{\partial\boldsymbol{\phi}_{{c}_{ii}}^{w}}
		=\frac{\partial(||[\boldsymbol{t}_{{c}_{j}}^{{c}_{k}}\times]\boldsymbol{p}_{{c}_{k}}||)}{\partial\boldsymbol{t}_{{c}_{j}}^{{c}_{k}}} \cdot 
		\frac{\partial\boldsymbol{t}_{{c}_{j}}^{{c}_{k}}}{\partial\delta\boldsymbol{p}_{{c}_{ii}}^{w}} \\
		&=\begin{cases}
			ii=j:(\frac{{p}_{{c}_{k}}^{T}{p}_{{c}_{k}}(\boldsymbol{t}_{{c}_{k}})^{T}-{p}_{{c}_{k}}^{T}\boldsymbol{t}_{{c}_{k}}{p}_{{c}_{k}}^{T}}{||[\boldsymbol{t}_{{c}_{j}}^{{c}_{k}}\times]\boldsymbol{p}_{{c}_{k}}||}) 
			\cdot \boldsymbol{R}_{w}^{{c}_{k}}\\
			ii=k:-(\frac{{p}_{{c}_{k}}^{T}{p}_{{c}_{k}}(\boldsymbol{t}_{{c}_{k}})^{T}-{p}_{{c}_{k}}^{T}\boldsymbol{t}_{{c}_{k}}{p}_{{c}_{k}}^{T}}{||[\boldsymbol{t}_{{c}_{j}}^{{c}_{k}}\times]\boldsymbol{p}_{{c}_{k}}||}) 
			\cdot \boldsymbol{R}_{w}^{{c}_{k}}\\
			ii=others:0
		\end{cases}
	\end{aligned}
\end{equation}

\begin{equation}\label{eqn:Hxi pos PO F}
	\begin{aligned}
		F&=\frac{\partial(||[\boldsymbol{p}_{{c}_{k}}\times]\boldsymbol{R}_{{c}_{j}}^{{c}_{k}}\boldsymbol{p}_{{c}_{j}}||\boldsymbol{t}_{{c}_{j}}^{{c}_{i}})}{\partial\delta\boldsymbol{p}_{{c}_{ii}}^{w}} 
		=||[\boldsymbol{p}_{{c}_{k}}\times]\boldsymbol{R}_{{c}_{j}}^{{c}_{k}}\boldsymbol{p}_{{c}_{j}}||
		\frac{\partial\boldsymbol{t}_{{c}_{j}}^{{c}_{i}}}{\partial\delta\boldsymbol{p}_{{c}_{ii}}^{w}} \\
		&=\begin{cases}
			i=j,ii=any:0 \\
			i \neq j:\begin{cases}
						ii=j:||[\boldsymbol{p}_{{c}_{k}}\times]\boldsymbol{R}_{{c}_{j}}^{{c}_{k}}\boldsymbol{p}_{{c}_{j}}||\boldsymbol{R}_{w}^{{c}_{i}}\\
						ii=i:-||[\boldsymbol{p}_{{c}_{k}}\times]\boldsymbol{R}_{{c}_{j}}^{{c}_{k}}\boldsymbol{p}_{{c}_{j}}||\boldsymbol{R}_{w}^{{c}_{i}}\\
						ii=others:0
					 \end{cases}
			\end{cases}
	\end{aligned}
\end{equation}

Compared to MSCKF \cite{4209642}, the reprojection residuals in \eqref{eqn:reproject PO} are solely related to the motion states and thus overcome the requirements of space projection; While in \eqref{eqn:Hxi PO}-\eqref{eqn:Hxi pos PO F}, PO-MSCKF does not require any feature position information, which removes the computational cost and linearization errors brought in by the 3D reconstruction procedure.

\section{Experiment Setup}\label{sec:Experiment Setup}
This section provides the details of the datasets and the experimental procedures, where the proposed algorithm and some state-of-the-art (SOTA) VIO algorithms have been evaluated on benchmark datasets.

\subsection{Experimental Condition}\label{sec:Experimental Condition}
\subsubsection{EuRoc Datasets}\label{sec:EuRoc Dataset1}
EuRoc datasets \cite{WOS:000382981300001} were collected by a micro-aerial vehicle (MAV), which contain 20 Hz stereo images, 200 Hz IMU datas, and ground truth.

In the experiments, we used the machine hall data sequences, where MH01 and MH02 were categorized as easy, MH03 was categorized as medium, MH04 and MH05 were categorized as difficult. 

\subsubsection{Kitti Datasets}\label{sec:Kitti Datasets1}
Kitti datasets \cite{WOS:000309166203066} were collected by a land vehicle experimental platform, including 10 Hz images, 200 Hz IMU datas, and ground truth. 
We used three challenging sequences 08, 09, and 10 for testing. The sequences include urban and highway environment data captured at different vehicle dynamics.

\subsubsection{NUDT Datasets}\label{sec:NUDT Datasets1}
\begin{figure}[htbp]
	\centering
	\subfigure[] { \begin{minipage}{4cm} \centering \includegraphics [height=3.5cm]{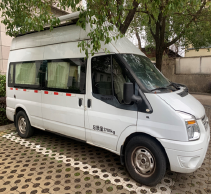}\label{fig:carplat} \end{minipage} } 
	\subfigure[] { \begin{minipage}{4cm} \centering \includegraphics [height=3.5cm]{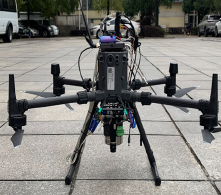}\label{fig:uavplat} \end{minipage} }
	\caption{
		Experimental platform in Real-World experiments.
		(a) Landvehicle experiment.
		(b) UAV experiment.
	}
	\label{fig:experimental platform}
\end{figure}

NUDT datasets include a sequence of UAV flight data and a sequence of land vehicle experimental data.
The UAV experiment was carried out at an altitude of about 180m above the ground, with a total distance of about 2.8km. The total distance of the land vehicle experiment was about 4.5km.
The experimental platform for the NUDT dataset is shown in \figref{fig:experimental platform}. The main parameters of the onboard sensors are shown in \tabref{tab:Navigation Sensors}. 

\begin{table}[htbp]
	\begin{center}
		\caption{Navigation Sensors Specifications}
		\label{tab:Navigation Sensors}
		\begin{tabular}{*{3}{c}}
			\toprule
			Sequence & Parameter & Frequency \\			
			\midrule
			\multirow{3}{*}{Land vehicle} & Gyroscope bias: $0.5 ^{\circ}/h$  & \multirow{2}{*}{$200\mathrm{Hz}$} \\
			\multirow{3}{*}{} & Accelerometer bias: $0.15 mg$  & \multirow{2}{*}{} \\
			\multirow{3}{*}{} & Camera resolution: $1620 \times 1220$  & $20\mathrm{Hz}$ \\
			\multirow{3}{*}{UAV} & Gyroscope bias: $0.5 ^{\circ}/h$  & \multirow{2}{*}{$200\mathrm{Hz}$} \\
			\multirow{3}{*}{} & Accelerometer bias: $0.1 mg$  & \multirow{2}{*}{} \\
			\multirow{3}{*}{} & Camera resolution: $720 \times 540$  & $10\mathrm{Hz}$ \\
			\bottomrule
		\end{tabular}
	\end{center}
\end{table}

\subsection{Experimental Procedure}\label{sec:Experimental Procedure}
\subsubsection{Comparison Algorithms}\label{sec:Comparison Algorithm}
The comparison algorithms include OpenVINS \cite{9196524}, VINS-Mono \cite{WOS:000442341000003} and OKVIS \cite{WOS:000350472800005}; These algorithms are currently the SOTA VIO works, where OpenVINS is a filter-based algorithm while VINS-Mono and OKVIS are optimization-based algorithms.

\subsubsection{Algorithms Setup}\label{sec:Algorithms Setup}
In the Euroc datasets, OpenVINS ran with default parameters provided by \cite{9196524}, while the results of VINS-Mono and OKVIS came from \cite{WOS:000442341000003}. 

On the other two datasets, all algorithms used consistent parameter settings.

In feature detection and tracking, OpenVINS and VINS-Mono used the default settings, while PO-MSCKF used fast corners as features \cite{WOS:000271826700009} and KLT method \cite{10.5555/1623264.1623280} for tracking.
In terms of initialization, OpenVINS and VINS-Mono also used the default settings and have successfully initialized on all the testing sequences. 
The focus of this research is on building a more efficient and robust VIO algorithm rather than initialization work; therefore, we have used ground-truth for initialization on the public datasets and a two-position alignment algorithm \cite{WOS:000462540400144} for initialization on the NUDT datasets.

All the testing algorithms has disabled the loop closure component in the experiments for fair comparisons.

\section{Result}\label{sec:Result}
In this section, we used the EVO toolbox to perform trajectory comparison and quantitative evaluation of all results \cite{WOS:000458872706092}. 

\subsection{EuRoc Datasets}\label{sec:EuRoc Dataset}

\tabref{tab:Euroc RMSE} lists the root mean square error (RMSE) of the position estimates for all sequences in EuRoc datasets. 
As shown in the table, PO-MSCKF achieves the best performance in three of the sequences, including the challenging sequence MH05.

In indoor small-depth environments such as the EuRoc datasets, the depth of features is smaller and the accuracy of 3D reconstruction is higher.
Therefore, PO-MSCKF cannot fully leverage the advantage of being independent of the features' depth, resulting in overall performance comparable to other advanced algorithms.
However, PO-MSCKF has achieved the best performance in the challenging sequence MH05, which proves its robustness.

\begin{table}[htbp]
	\begin{center}
		\setlength{\tabcolsep}{5pt}
		\caption{RMSE in Euroc Datasets in Meters}
		\label{tab:Euroc RMSE}
		\begin{tabular}{*{6}{c}}
			\toprule
			Sequence & Distance & OpenVINS & PO-MSCKF & VINS-Mono & OKVIS \\			
			\midrule
			MH01 & $80$ & $\boldsymbol{0.11}$ & $\boldsymbol{0.11}$ & $0.15$ & $0.33$ \\
			MH02 & $73$ & $\boldsymbol{0.13}$ & $0.30$ & $0.15$ & $0.37$ \\
			MH03 & $131$ & $0.16$ & $\boldsymbol{0.15}$ & $0.22$ & $0.25$ \\
			MH04 & $92$ & $\boldsymbol{0.17}$ & $0.28$ & $0.32$ & $0.27$ \\
			MH05 & $98$ & $0.44$ & $\boldsymbol{0.28}$ & $0.30$ & $0.39$ \\
			\bottomrule
		\end{tabular}
	\end{center}
\end{table}

\subsection{Kitti Datasets}\label{sec:Kitti Dataset}

\begin{figure}[htbp]
	\centering
	\subfigure[] { \begin{minipage}{4cm} \centering \includegraphics {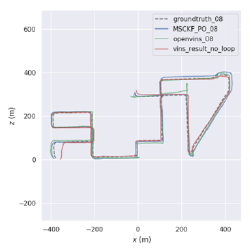}\label{fig:Kitti08} \end{minipage} } 
	\subfigure[] { \begin{minipage}{4cm} \centering \includegraphics {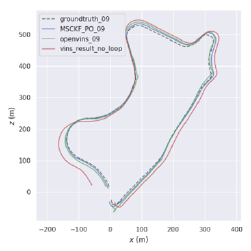}\label{fig:Kitti09} \end{minipage} }
	\subfigure[] { \begin{minipage}{4cm} \centering \includegraphics [height=4.5cm]{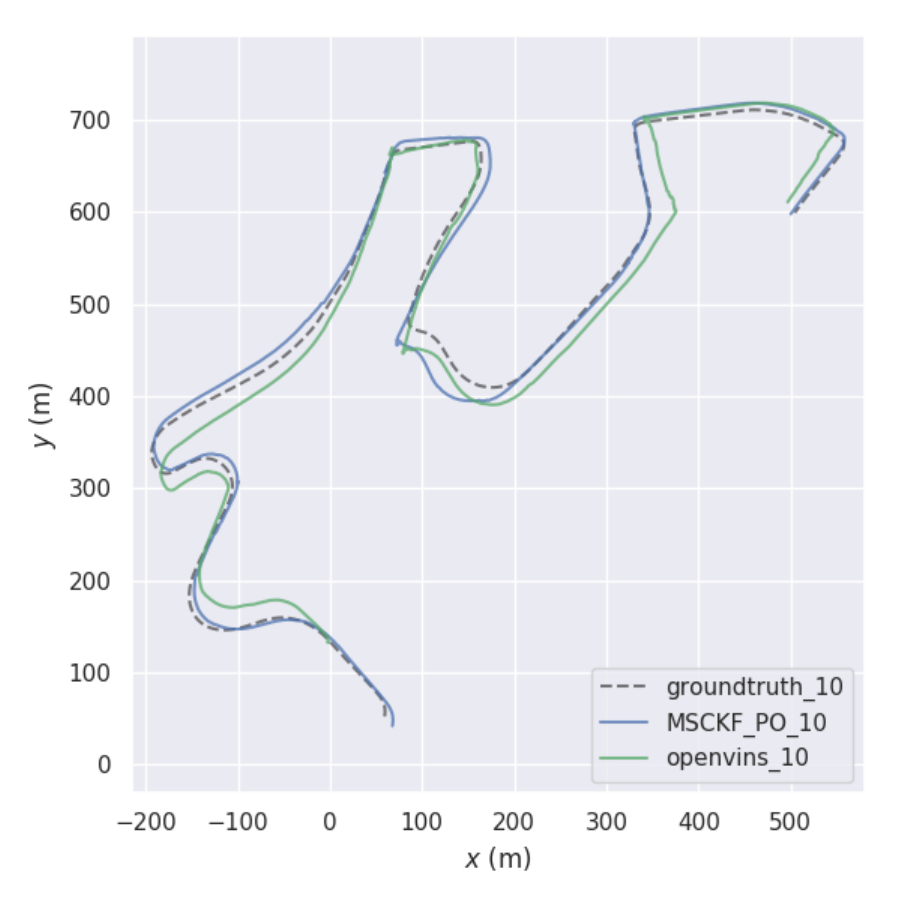}\label{fig:Kitti_10} \end{minipage} } 
	\caption{
		Trajectory in Kitti datasets, compared with OpenVINS and VINS-Mono.
		(a) Kitti08.
		(b) Kitti09.
		(c) Kitti10.
	}
	\label{fig:Kittidatasets}
\end{figure}

The comparison of the trajectories are shown in \figref{fig:Kittidatasets}, and the RMSE is shown in \tabref{tab:Kitti RMSE}.

As shown in the figure and table, PO-MSCKF achieves the best performance in all sequences.
In sequence 10, the continuous fast turns resulted in a decrease in the estimation accuracy of VINS-Mono and OpenVINS, especially causing the former to be unable to complete this sequence smoothly.

In outdoor large-depth environments such as the Kitti datasets, as the features' depth increases, the accuracy of 3D reconstruction begins to decrease; therefore, the linearization error of VIO algorithms based on traditional multi-view geometry descriptions will increase.
Meanwhile, the proposed algorithm is independent of 3D features, therefore it can achieve improvements in accuracy and robustness in outdoor large-depth environments.

\begin{table}[htbp]
	\begin{center}
		\caption{RMSE in Kitti Datasets in Meters}
		\label{tab:Kitti RMSE}
		\begin{tabular}{*{5}{c}}
			\toprule
			Sequence & Distance & VINS-Mono & OpenVINS & PO-MSCKF \\			
			\midrule
			Kitti08 & $2823.14$ & $11.84$ & $16.85$ & $\boldsymbol{9.17}$ \\
			Kitti09 & $1653.84$ & $21.40$ & $13.17$ & $\boldsymbol{9.74}$ \\
			Kitti10 & $2107.63$ & $--$ & $30.28$ & $\boldsymbol{14.01}$ \\
			\bottomrule
		\end{tabular}
	\end{center}
\end{table}

\subsection{NUDT Datasets}\label{sec:NUDT Datasets}

\begin{figure}[htbp]
	\centering
	\subfigure[] { \begin{minipage}{4cm} \centering \includegraphics {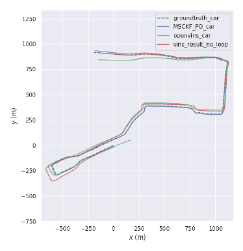}\label{fig:car} \end{minipage} } 
	\subfigure[] { \begin{minipage}{4cm} \centering \includegraphics {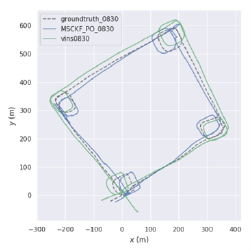}\label{fig:uav} \end{minipage} }
	\caption{
		Trajectory in Real-World experiments, compared with OpenVINS and VINS-Mono.
		(a) Landvehicle experiment.
		(b) UAV experiment.
	}
	\label{fig:Real-World}
\end{figure}

\begin{figure}[htbp]
	\centering
	\includegraphics [width=6cm,height=5cm]{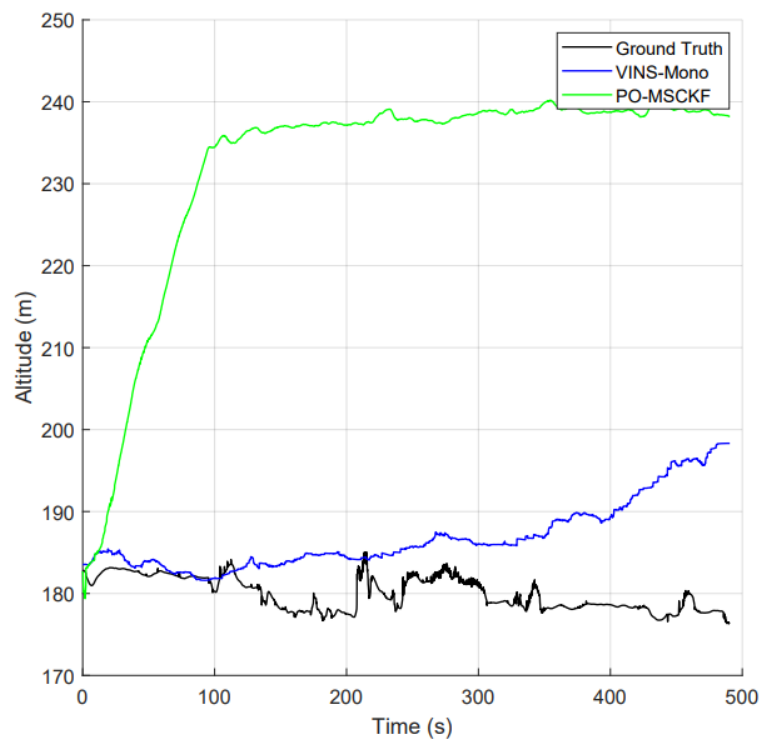}\label{fig:uav-alti}  
	\caption{
		The altitude trajectory of UAV experiments, compared with VINS-Mono.
	}
	\label{fig:UAV-Alti}
\end{figure}

The comparison of trajectory is shown in \figref{fig:Real-World}, and RMSE is shown in \tabref{tab:Realworld RMSE}. 
In addition, due to the reduced accuracy of the 3D reconstruction process in large-depth environments, the altitude estimation is prone to divergence when flying at higher altitudes.
PO-MSCKF is independent of the feature's depth and theoretically can better maintain the accuracy of altitude estimation.
To verify this inference, we added a comparison of altitude estimation in the UAV experiment as shown in \figref{fig:UAV-Alti} .

As shown in the figures and table, PO-MSCKF still achieves the best performance in all sequences.
In the UAV experiment, the large-depth environment makes it difficult for OpenVINS to accurately complete 3D reconstruction, which introduces a large number of linearization errors and leads to experimental failure; for VINS-Mono, the large-depth environment also leads to severe divergence in its altitude estimation.
Compared with the other algorithms, PO-MSCKF further demonstrates the advantages of decoupling from 3D features and has more efficient and robust performance in large-depth environments.

\begin{table}[htbp]
	\begin{center}
		\caption{RMSE in Real World Experiments in Meters}
		\label{tab:Realworld RMSE}
		\begin{tabular}{*{5}{c}}
			\toprule
			Sequence & Distance & VINS-Mono & OpenVINS & PO-MSCKF \\			
			\midrule
			CAR & $4483.54$ & $36.38$ & $49.97$ & $\boldsymbol{10.69}$ \\
			UAV & $2793.57$ & $27.77$ & $--$ & $\boldsymbol{25.47}$ \\
			\bottomrule
		\end{tabular}
	\end{center}
\end{table}

\section{Conclusion and Discussion}\label{sec:Conclusion}

In this paper, we propose an efficient VIO system: PO-MSCKF. 

Our algorithm has the following advantages:
1) Having a concise state space that only includes motion states.
2) Eliminates the null space projection process of traditional MSCKF and has better model consistency.
3) Removed the 3D feature reconstruction process, avoiding the impact of reconstruction errors.
Benchmark experiments have shown that PO-MSCKF can perform better than the compared SOTA algorithms on most sequences.

There are still many methods to improve performance that have not been used, such as observability constraints and time delay compensation, but PO-MSCKF still shows its progressiveness in benchmark experiments.
We will further improve our algorithm in the future and consider its application in computationally constrained miniaturization platforms.

\section*{Acknowledgment}
The authors would like to thanks Chenjun Ji and Ruiguang He for their help in the experiments.

\bibliographystyle{IEEEtran}

\bibliography{Ref}

\end{document}